\documentclass[runningheads]{llncs}
\usepackage{url}

\usepackage{graphicx}
\usepackage[dvipsnames]{xcolor}

\usepackage[colorinlistoftodos,prependcaption,textsize=tiny]{todonotes}

\usepackage{placeins} 
\usepackage{flafter}  

\usepackage{float}
\usepackage{flafter}

\begin{document}
\title{Detecting Group Beliefs Related to 2018's Brazilian Elections in Tweets: A Combined Study on Modeling Topics and Sentiment Analysis}
%
\titlerunning{Detecting Group Beliefs Related to 2018's Brazilian Elections in Tweets} 
%
\author{Brenda Salenave  Santana\inst{1}\orcidID{0000-0002-4853-5966} \and
Aline Aver Vanin\inst{2}\orcidID{0000-0002-9984-6043}}

\authorrunning{B.S., Santana. and A.A., Vanin}
%

\institute{Institute of Informatics, Federal University of Rio Grande do Sul, Porto Alegre, Brazil \\ \email{bssantana@inf.ufrgs.br} \and
Departament of Education and Humanities, Federal University of Health Sciences of Porto Alegre, Porto Alegre, Brazil \\
\email{alinevanin@ufcspa.edu.br}}
\maketitle              
\begin{abstract}
2018's Brazilian presidential elections highlighted the influence of alternative media and social networks, such as Twitter.
In this work, we perform an analysis covering politically motivated discourses related to the second round in Brazilian elections.
In order to verify whether similar discourses reinforce group engagement to personal beliefs, we collected a set of tweets related to political hashtags at that moment. To this end, we have used a combination of topic modeling approach with opinion mining techniques to analyze the motivated political discourses.
Using SentiLex-PT, a Portuguese sentiment lexicon, we extracted from the dataset the top 5 most frequent group of words related to opinions. Applying a bag-of-words model, the cosine similarity calculation was performed between each opinion and the observed groups. 
This study allowed us to observe an exacerbated use of passionate discourses in the digital political scenario as a form of appreciation and engagement to the groups which convey similar beliefs.
\end{abstract}
 {\let\thefootnote\relax\footnotetext{Copyright \textcopyright\ 2020 for this paper by its authors. Use permitted under Creative Commons License Attribution 4.0 International (CC BY 4.0). DHandNLP, 2 March 2020, Evora, Portugal. }}   
 
\section{Introduction}


In 2016 the Oxford Dictionaries \cite{oxford2016oup} chose post-truth as the word of the year. It is an adjective used for `relating to or denoting circumstances in which objective facts are less influential in shaping public opinion than appeals to emotion and personal belief'. 
From this perspective, the Internet has been proving to be a place where beliefs meet. Due to algorithms created to meet the users’ expectations, much of what is publicized is part of a bubble of information.

There is a variety of ways in which the Internet makes it more likely that individuals with shared views or preferences cluster together \cite{doi:10.1146/annurev-polisci-030810-110815}. This phenomenon may happen with the user consent, when they search for groups chosen by preferences, to the
application of recommending systems designed to connect users with similar preferences intentionally.

This kind of phenomenon opens up room for the circulation of news designed to provide emotional appeal, thereby increasing interaction with the content \cite{recuero2017midia}.
In periods of electoral dispute, political opinion engagements become fiercer, and the filter bubble effect further encapsulates affections by reinforcing polarization.
Studies aimed at understanding the consequences of internet use for politics (\cite{doi:10.1146/annurev-polisci-030810-110815}, \cite{allcott2017social}), and among these uses is the massive spreading of spurious news, as happened in the 2016 US elections \cite{d2017post}.

Different social networks have been working to identify fake news and warn users. 
It is pointed out that these actions may increase social welfare \cite{allcott2017social}, but identifying fake news sites and articles also raises important questions about who becomes the arbiter of truth. Notice that the more an opinion meets one's expectation, the more it is likely to be approved and shared.

Virtual environments provide preeminent involvement of people in the context of political discussions, enabling discussions, presenting different points of view on the same topic.
On Twitter, for example, opinion leadership tends to increase users' engagement in political processes \cite{PARK20131641}.
Online interaction, even through hashtags, with influential users, then serves as a tool for reinforcing beliefs by the actors involved.
However, in a post-truth context were opinions are manifested and widespread in social networks and groups, produced discourses obtain the power to manipulate people's opinion and legitimize non-socially accepted actions and behaviors in a different contexts.

This paper aims to analyze how the discourses produced on Twitter during the second round of the electoral race in Brazil reinforce the group engagement to personal beliefs. 
We use a combination of topic modeling approach with opinion mining techniques to analyze motivated political text discourses produced.

This paper is organized as follows:
Section \ref{sec:twitter-electoral} describes a background of the use of social networks, more specifically Twitter, in politics context and the discourses produced in virtual environments.
The following of this section then presents the applied approach and the discussion of results. 
Finally, the study is concluded in Section \ref{sec:final}, with some closing remarks and an overview of the observed insights.

\section{Use of Social Networks as an Electoral Tool}\label{sec:twitter-electoral}

Brazil's political context is complex, as the country is undergoing a social, economic, and political crisis that has worsened in recent years.
Such circumstances, according to \cite{RECUERO2019} were reflected in discussions that took place in both traditional and social media, and in particular, the spread of a great deal of misinformation and fake news about facts often disseminated by dubious sources.

Much of the discursive resources used as the basis of electoral campaigns in Brazil were the announcements of the so-called `television time', which are distributed according to the size of parties and their coalitions. From the observation of studies that correlate television advertising time and election results \cite{BORBA2017} \cite{10.1093/restud/rdq012}, it is then perceived that the assumption that radio and television election time is more critical in disputes farther from such as the presidential elections.
With the rise of digital media in increasingly significant roles for political communication, transforming the dynamics of information production, circulation, and consumption, the phase of free electoral political time on television has taken a back seat \cite{internetlab-custoreport}.

The highlight gained by digital media also has an active component of spontaneous and organic voter engagement, revealing that the role of digital media in electoral processes transcends paid advertising and communication controlled directly by the campaigns \cite{internetlab-custoreport}.
Due to the division of television times between candidates for the presidency of the republic in Brazil’s 2018 presidential elections, some campaigns had its primary focus on alternative media to the then traditional television campaign, such as Twitter.

Twitter might be described as a real­time, highly social microblogging service that allows users to post short status updates, called tweets, that appear on timelines \cite{russell2018mining}. Tweets may include one or more entities in their 280 characters of content and reference one or more places that map to locations in the real world being a piece of valuable infrastructure that enables rapid and easy communication. 
All these characteristics make Twitter a source of fast dissemination of different types of content.

However, in 2018, 37.9\% of all internet traffic was not human \cite{badbotreport}.
One of the agents in this scenario is the so-called twitter-bots. 
These are defined as “small software programs that are designed to mimic human tweets" \cite{akimoto2011japan}. 
Still, according to the author, some bots reply to other users when they detect specific keywords, while others may randomly tweet preset phrases, such as proverbs. Not all bots are fully machine-generated. However, the “bot” term has also come to refer to Twitter accounts that are fake.

Authors such as \cite{kumar2018false} point to the influence of very engaged bots and actors who can create false consensus perceptions for specific information to circulate.
Thus, for \cite{RECUERO2019}, bots operate to (1) rapidly increase the visibility of false information and (2) to inflate the status of some users, making certain false information credible. These actions enhance the dissemination of fake news, in an attempt to influence the public sphere, with the artificial manipulation of consensus and, likewise, public opinion.

However, it is often the case that the discourse disseminated in social networks makes use of different types of language resources that reinforces social and political conditions that can encourage conditions such as verbal violence.
In this scenario, passionate political discourses or even bots programmed to interact with mentioned keywords can improve the dissemination of beliefs in the political sphere. In this context, this works intends to study the content of tweets related to Brazilian presidential election through hashtags identified in trending topics in order to analyze the discursive similarity that brings users closer.

In the next section, we describe the extraction and preprocessing of tweets containing political opinions during the week before the Brazilian electoral day.  

\subsection{Data Extraction and Preprocessing}
Given the focus of the campaign through social networks, we opted to analyze the discourses related to the presidential election through hashtags\footnote{https://help.twitter.com/en/using-twitter/how-to-use-hashtags} identified in Twitter trending topics in Brazil. Thus, in conducting the present study, 50.000 tweets were collected per day during the last week of the election campaign, from October 22nd to 28th, through the Twitter API. It is noteworthy that the collection of information for each day was performed on the following day, regardless of geographical location, aiming to extract texts that made use of one or more listed hashtags.
Most of the texts found were written in Portuguese, however tweets in English were also detected.

Table \ref{tab:hashtags} presents the leading hashtags listed, in Portuguese, observed initially. For better understanding, we provide their translations when necessary.

\begin{table}[H]
\centering
\caption{Leading Twitter's Hashtag Analyzed.}\label{tab:hashtags}
\begin{tabular}{|l|l|}
\hline
\multicolumn{1}{|c|}{\begin{tabular}[c]{@{}c@{}}Portuguese Used Hashtags \\ (original)\end{tabular}} & \multicolumn{1}{|c|}{\begin{tabular}[c]{@{}c@{}}Translated Hashtags Used\\ (English version)\end{tabular}}\\
\hline
\#B17 &  -\\
\#Bolsonaro17 & -\\
\#BolsonaroPresidente & \#BolsonaroPresident\\
\#EleSim &  \#HeYes\\
\#Haddadnãoecristao & \#HaddadIsNotChristian\\
\#NasruascomBolsonaro & \#InTheStreetsWithBolsonaro\\
\hline

\end{tabular}
\end{table}

For a better understanding of the data, a preprocessing step was performed.
Dataset normalization covered such things as tokenization, uppercase and lowercase transformation, removal of special characters and image and video tags, among others. 
The sentences of each tweet were then tokenized, that is, separated into different units.
At this stage, the Portuguese and English stopwords contained in these units of texts were removed.

Next, the standardization of terms is performed with the application of the lemmatization technique employed with the aid of text processing libraries\footnote{https://www.nltk.org/}$^{,}$\footnote{https://spacy.io/}.
The lemmatization process seeks to reduce the word to its root form, that is, its standard form of singular writing.
At this stage of execution, we then sought to normalize the observed texts in order to keep the most relevant words of the tweets without losing the initial semantics.
For this purpose, we also use the Word2Vec approach, whose idea is to transform each token of the set of sentences (a.k.a. corpus) into a numerical vector that semantically represents it \cite{mikolov2013distributed}.
The data model was then trained using the skip-gram model approach, since the distributed representation of the input word is used to predict the context.

\begin{figure}[!htbp]
\centering
\includegraphics[scale=.3]{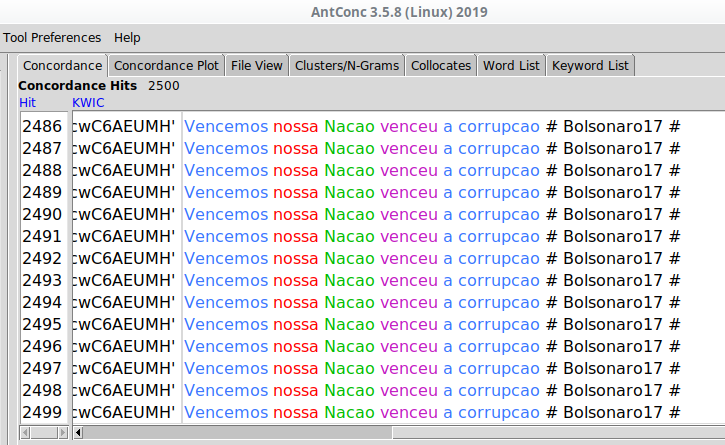}
\caption{Example of repetitions found on analyzed dataset.} 
\label{fig:screenshot}
\end{figure}

Using a multi-purpose toolkit for corpus analysis \cite{antconc2019}, it was possible to identify several repetitions of posts made by the same users, as showed in Figure \ref{fig:screenshot}.
Some of these posts had up to 2500 repetitions, which was considered a very high rate for normal user behavior.
For a better analysis of the observed texts, it was necessary to remove the repetitions of tweets. The removal of such repetitions proved to be of high relevance as it was considered as a possible indication of discourses spread and corroborated by twitter-bots.
Currently, over 18\% of the accounts associated with the collected tweets no longer exist.

\subsection{Modeling Topics in Politically Motivated Digital Discourses}
Human discourse is a tool for daily communication, but it is not easily read in machine structures. Human language has unique nuances that occur in specific contexts, and this kind of information needs to be computationally translated to be read by machines in order to be processed.   
Topic modeling allows observed sets of observations to be explained by unobserved groups that seek to explain why some parts of the data are similar.

From a broad perspective, topic modeling methods based on Latent Dirichlet Allocation (LDA) have been applied to natural language processing, text mining, and social media analysis, information retrieval \cite{Jelodar2019}.
In this sense, topic models allow exploring concepts and meanings extracted from the use of word embeddings.
LDA is a fully generative model for describing the latent topics of documents \cite{7050858}.
Since observations are made about words collected from textual files (documents), in this approach each document can be interpreted as a mixture of various topics in which each document is considered to have a set of topics that are assigned to it through LDA \cite{7050858}.

Sentiment polarities are dependent on topics or domains \cite{Lin:2009:JSM:1645953.1646003}. In this way, in our work, we drive an analysis covering politically motivated discourses in the Twitter environment.  Observing different texts, we then focus on identifying polarities expressed in observed online interactions by correlating them to the topics of previously identified groups.


Table \ref{tab:lda-topics} presents the main words highlighted in the LDA topics.
Five different topics were listed from the observed set, reaching a coherence score of 0.58 with the c\_v measure.
The c\_v topic coherence captures the optimal number of topics by giving the interpretability of these topics a number called coherence score. We use the c\_v measure. This measure is based on a sliding window, one-set segmentation of the most frequent, and an indirect confirmation measure that uses normalized pointwise mutual information (NPMI) and the cosine similarity (a metric used to determine how similar the documents are irrespective of their size) \cite{huang2008similarity}.
The obtained score represents a low coherence between some terms observed in the dataset, i.e., the occurrence of non-cohesive windows of words.
 
\FloatBarrier
\begin{table}[ht!]
\caption{Main Words Highlighted by LDA Topics.}
\centering
\resizebox{\textwidth-7mm}{!}{%
\begin{tabular}{|l|l|l|l|l|}
\hline
\multicolumn{1}{|c|}{\textbf{Group 1}} & \multicolumn{1}{c|}{\textbf{Group 2}} & \multicolumn{1}{c|}{\textbf{Group 3}} & \multicolumn{1}{c|}{\textbf{Group 4}} & \multicolumn{1}{c|}{\textbf{Group 5}} \\ \hline
jairbolsonaro & haddadnaoecristao & bolsonaro & bolsonaropresidente & bolsonaropresidente \\
haddadnaoecristao & brasileiro & bolsonaropresidente17 & brasil & estao \\
fake & ver & haddad & b17 & mulherescombolsonaro \\
voto & caro & delegadofrancischini & acima & conseguir \\
news & bolsonaroanticristo & torturar & bolsonaro17 & candidatar \\
olhar & torturar & nao & ficar & verdade \\
votar & esquecer & perder & think & passar \\
elesim & melhor & compartilhar & bolsonaropresidenteeleito & voce \\
prefeito & senhor & elesim & mudabrasil17 & filho \\
horario & usar & forapt & ptnuncamais & video \\ \hline
\end{tabular}
}
\label{tab:lda-topics}
\end{table}

\begin{figure}[ht!]
\centering
\includegraphics[width=\textwidth-10mm]{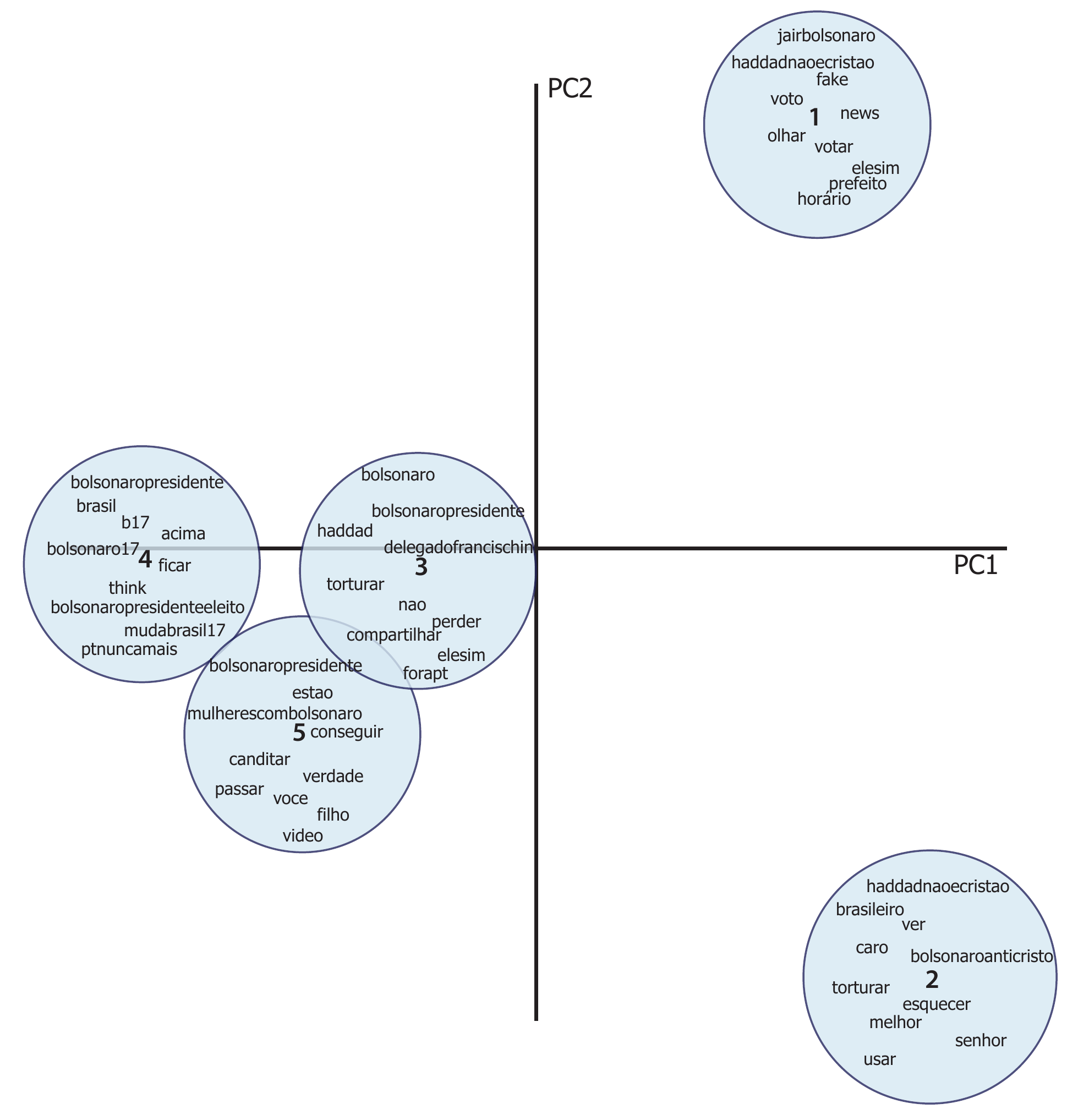}
\caption{Intertopic Distance Map (via multidimensional scaling)} 
\label{fig:dist-map}
\end{figure}

Figure \ref{fig:dist-map} presents the identified groups, using an intertopic distance map, via multidimensional scaling. The group topics are exhibited in terms of principal components axis distribution.
This is a statistical procedure that uses an orthogonal transformation to convert a set of observations of possibly correlated variables into a set of values of linearly uncorrelated variables called principal components (PC) \cite{jolliffe2016principal}.
In this figure, it is possible to see how the topic groups are organized in this space and the top 10 topics that compose them.
The topics identified are then plotted as circles, whose centers are defined by the computed distance between topics (projected into two dimensions). The prevalence of each topic is indicated by the circle’s area \cite{sievert-shirley-2014-ldavis}. 

\subsection{Opinion mining on election-related tweets}
After extracting the sets of observation of the data, that is, LDA topics, we sought to extract the main opinions expressed in the observed texts.
In order to analyze the polarities manifested in the tweets contained in the comments, we used SentiLex-PT, a sentiment lexicon designed for the extraction of sentiment and opinion about human entities in Portuguese texts \cite{carvalho2015sentilex}.

The extraction of observed sentiments was performed to capture expressions and words previously classified in the lexicon as positive or negative. The proposed sentiment analysis encompassed a complete observation of the dataset seeking to explore all mentions of expressions that could refer to the polarity of the content as a whole, also relating the similarity of words in the same context.


Using the Word2Vec approach, we sought to observe the polarities associated with each topic.
Analyzing the main words that made up the topics identified above and, with the crossing of data extracted from the lexicon, a cosine similarity \cite{huang2008similarity} calculation was performed over the skip-gram model.
The similarity extracted from the content of the observed data is a verification measure of how close two terms are in the given context.
In order to observe how the feelings manifested interact in the modeled topics, only the expressions categorized in the lexicon either as positive or negative were used, discarding the neutral words.

\begin{figure}[!htbp]
\centering
\includegraphics[width=\textwidth+10mm]{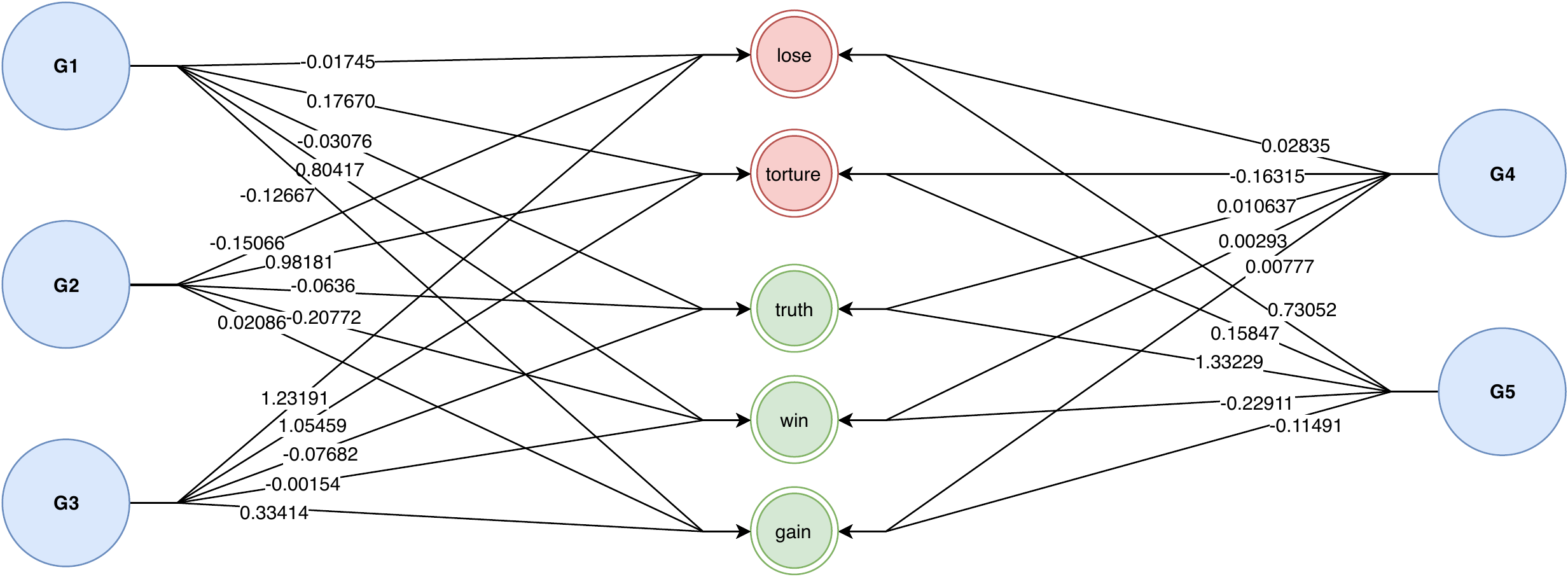}
\caption{Relation between topic groups and the most prevalent sentiments.} 
\label{fig:topics-dist}
\end{figure}

Using SentiLex,  we search for the five most frequent non-neutral words in the observed set, either they are classified as positive or negative. As a result of this process, the following words were then listed: \textit{perder} (lose), \textit{torturar} (torture), \textit{verdade} (truth), \textit{vencer} (win) and, \textit{ganhar} (gain).
Using the bag-of-words model, a cosine similarity calculation was performed between each opinion and the observed groups. From this, it then extracted the average correspondence between the words in the analyzed context.
Figure \ref{fig:topics-dist} presents the relation of each LDA identified topic with the main opinions presented in the data.

Analyzing the relation between topic groups (see Table 2) and the identified opinions, it is possible to observe that for groups 1, 4, and 5. The most prevalent relations happen with the words of positive polarity, i.e., with the sentiments related to the act of winning. The inverse occurs to groups 2 and 3. Checking the relations with the topic groups, it is possible to observe that for group 4: all the relations are weaker than the ones in other groups. For the others, relations seem to be more frequent.



For the observed groups, the higher relation occurs like: group 1 $\leftrightarrow$ win, group 2 $\leftrightarrow$ torture, group 3 $\leftrightarrow$ lose, and groups 4 e 5 $\leftrightarrow$ truth.
Assuming that the high correlation in the group 2 occurred since it was presented among the most relevant of the topic, another similarity check was made, this time disregarding the use of the word in the topic. Still, the correlation of the topic with the feeling remains superior to the others.  
It is important to note that how opposites are the words related to the opinions, not only in the polarity perspective but also in the semantic weight, especially in a political context.

Analyzing the similarity between the identified opinions and the named entity present in the dataset, sentiments of negative polarity have greater similarity to references of opposition, while positive polarity refers to the ideological parity of hashtags, that is, positive polarity words are related to different parties' named entities.
Thus, it is possible to observe the existence of an emotional appeal in the digital discourses of political motivation. This can be interpreted as a way of subjugating the other and extolling beliefs, leading to a reflection on the effects of post-truth on a polarized Brazilian politics.
These findings can lead to interpreting that from intersubjective relations, discourse, and the logic of recognition: the main feature of post-truth is that it requires a refusal of the other or, at least, a culture of indifference. When one is threatened, they react with hatred or violence, as indicated by \cite{subjetividade-posverdade:2019}.

These tendencies of emotional calling then tend to meet confirming biases leading to a higher propensity to remember, seek information, or interpret facts in ways that confirm pre-established beliefs.
The occurrence of the ambiguities of polarities found throughout the study presents a connection with the obtained coherence score. 
The low score found to be cohesive with the attribution of negative words in the context of the use of hashtags with a positive bias in search of an electoral victory. 

\section{Final Remarks} \label{sec:final}
This work proposed an analysis of how the discourses produced on Twitter related to the 2018 Brazilian second round of presidential elections reinforce the group engagement to personal beliefs. To this end, we have used a combination of topic modeling approach with opinion mining techniques to analyze motivated political text discourses produced on Twitter.

This study allowed us to observe an exacerbated use of passionate speech in the digital political scenario as a way of agreement to other's opinion. However, such use is not unattached from a strong correlation with negative opinion that is used for depreciation of others, such as the use of the word `torture', which is associated 
more strongly with the opposition candidate in the dispute.
In this sense, the reinforcement of beliefs is a strong ally in the sharing of information, consistent with pre-established confirmation bias.
The heavy use of repetitions suggest the presence of non-human interactions, and their oversharing throughout Twitter, might indicate credibility, leading to expressive engagement, since it is easier to agree with the majority's opinion. 

\section{Acknowledgments}
This study was financed in part by the Coordenação de Aperfeiçoamento de Pessoal de Nível Superior – Brasil (CAPES) – Finance Code 001.



%
%
%
\newpage
\bibliographystyle{splncs04}
\bibliography{bibliography}
\end{document}